%% file: iclr2026_conference.tex
\documentclass{article} 
\usepackage{iclr2026_conference,times}

\usepackage{microtype}
\usepackage{graphicx}
\usepackage{subcaption}
\usepackage{booktabs} 
\usepackage[HTML]{xcolor}
\usepackage[utf8]{inputenc}
\usepackage{hyperref}
\usepackage{amsmath}
\usepackage{amssymb}
\usepackage{mathtools}
\usepackage{amsthm}
\usepackage[most]{tcolorbox}

\usepackage[capitalize,noabbrev]{cleveref}

\usepackage{multirow}
\usepackage{multicol}

\input{math_commands.tex}

\usepackage{hyperref}
\usepackage{url}

\newcommand{\toknospace}[2]{%
  \begingroup
  \setbox0=\hbox{\texttt{Ag}}%
  \colorbox{#1}{%
    \raisebox{0pt}%
      [\ht0-0.15ex][\dp0-0.4ex]{\texttt{#2}}%
  }%
  \endgroup\!
}

\title{Infusion:\\Shaping Model Behavior by Editing Training Data via Influence Functions}

\author{
\And
J Rosser \\
FLAIR, University of Oxford \\
\texttt{jrosser@robots.ox.ac.uk}
\And
Robert Kirk \\
UK AI Security Institute
\And
\AND
Edward Grefenstette \\
AI Centre, UCL
\And
\And
\And
\And
\And
Jakob Foerster \\
FLAIR, University of Oxford
\And
\And
\And
\And
\And
Laura Ruis \\
MIT CSAIL
}

\iclrfinalcopy 
\begin{document}

\maketitle

\begin{abstract}

Influence functions are commonly used to attribute model behavior to training documents. We explore the reverse: crafting training data that induces model behavior. Our framework, \textsc{Infusion}, uses scalable influence-function approximations to compute small perturbations to training documents that induce targeted changes in model behavior through parameter shifts. We evaluate \textsc{Infusion} on data poisoning tasks across vision and language domains. On CIFAR-10, we show that making subtle edits via \textsc{Infusion} to just 0.2\% (100/45,000) of the training documents can be competitive with the baseline of inserting a small number of explicit behavior examples. We also find that \textsc{Infusion} transfers across architectures (ResNet $\leftrightarrow$ CNN), suggesting a single poisoned corpus can affect multiple independently trained models. In preliminary language experiments, we characterize when our approach increases the probability of target behaviors and when it fails, finding it most effective at amplifying behaviors the model has already learned. Taken together, these results show that small, subtle edits to training data can systematically shape model behavior, underscoring the importance of training data interpretability for adversaries and defenders alike. We provide the code here: \url{https://github.com/jrosseruk/infusion}.
\end{abstract}

\section{Introduction}

Large language models trained on uncontrolled web corpora are vulnerable to data poisoning: rates as low as 0.001\% can implant backdoors that persist through alignment \citep{zhang2024persistent}. Existing attacks often inject explicit instances of a target behavior into the training corpus \citep{zhang2024persistent, souly2025poisoning}. We ask whether a fundamentally different approach is possible: can an adversary make precise, minimal modifications to existing training documents that steer the model toward a targeted parameter state, \emph{without} explicitly demonstrating the target behavior? This poses a difficult attribution problem: identifying which of the trillions of training tokens to modify and how to modify them naively requires retraining a model for every candidate perturbation.

\begin{figure*}[h!]
    \centering
    \includegraphics[width=0.9\linewidth]{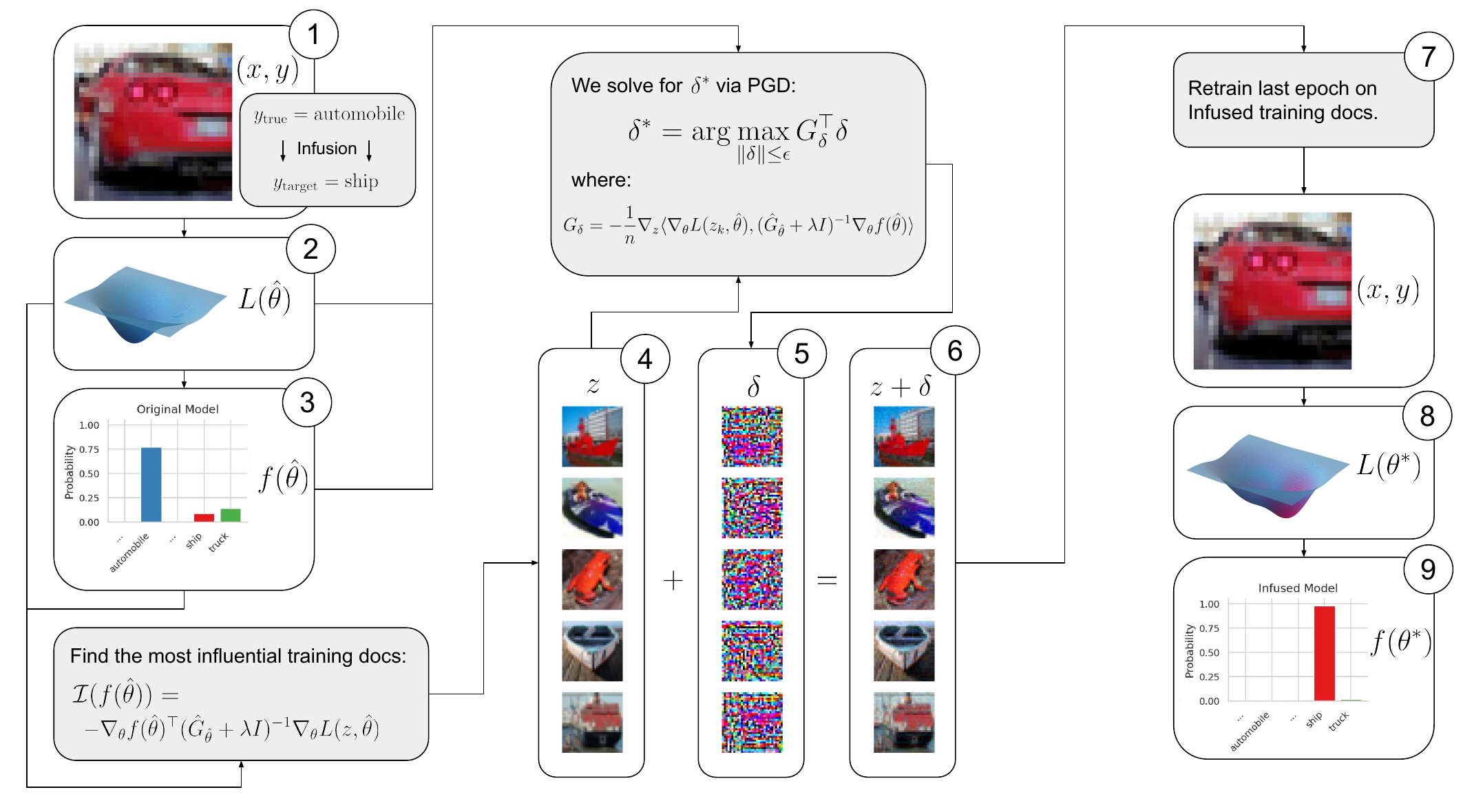}
    \caption{\textbf{The \textsc{Infusion} pipeline.} Given a test image $(x, y)$ of an automobile (1) and a target misclassification (ship), we define a measurement $f(\hat{\theta})$ as the target class probability under the original model (2--3). Using EK-FAC influence estimation, we identify the $k$ training examples most influential for this measurement (4). We then compute perturbations $\delta$ via projected gradient descent that maximize the predicted change in $f$ (5), yielding infused training examples $z + \delta$ (6). Retraining for one epoch on the modified corpus (7) produces a new model with shifted loss landscape $L(\theta^*)$ (8), where the target class probability has increased substantially on our test image whilst keeping all other model behavior nearly unchanged (9). Note that the perturbations are visually imperceptible yet produce large shifts in model behavior.}
    \label{fig:influence_figure1}
\end{figure*}
We introduce \textsc{Infusion}, a framework that leverages recent advances in scalable influence function estimation \citep{bae2022if, grosse2023studying} to (i) identify which training documents most affect a target behavior, (ii) compute gradient-based document perturbations that maximize adversarial objectives through their induced parameter changes, and (iii) validate predictions by retraining on the modified corpus. \textsc{Infusion}, shown in Figure \ref{fig:influence_figure1}, uses the extended influence function formulation for document-level perturbations and attacks from \citep{koh2017understanding}. Concretely, we formalize how replacing a document $z$ with a perturbed document $z+\delta$ induces a parameter shift
\begin{equation}
\Delta \hat{\theta}
\approx
-\frac{1}{n}\, H_{\hat{\theta}}^{-1}
\Big[\nabla_{z}\nabla_{\theta} L\!\left(z,\hat{\theta}\right)\Big]\delta,
\end{equation}
and how this shift changes a scalar measurement $f(\theta)$ representing a target behavior of interest via
\begin{equation}
\Delta f(\hat{\theta})
\approx
\nabla_{\theta} f(\hat{\theta})^{\top}\,\Delta \hat{\theta}.
\end{equation}

\break
\textbf{Our key contributions are listed as follows:}

\begin{itemize}
    \item \textbf{We introduce \textsc{Infusion}}, a framework that uses influence functions to identify which training documents most affect a target model's behavior and computes gradient-based perturbations that maximize an adversarial objective. We validate our framework on CIFAR-10, with all 2000 experiments successfully increasing the probability of the target behavior, and show that infused datasets can transfer attacks across architectures in both directions (ResNet $\leftrightarrow$ CNN).
    \item \textbf{We extend \textsc{Infusion} to pretrained language models in preliminary experiments.} We conduct initial experiments on language models by pretraining GPT-Neo on TinyStories, attempting to infuse a bias for one animal word over another. These experiments suffered from weaker influence estimates, smaller relative poisoning budgets, and a discrete optimization setting. We do find evidence that \textsc{Infusion} can increase the probably of target outputs, however it is rarely enough to change model behavior. On more structured Caesar Cipher encryption tasks, we find \textsc{Infusion} is most successful at amplifying latent model behavior.
    \item \textbf{Training data is a more critical attack surface than previously appreciated.}  While prior work shows that poisoning widely crawled sources can affect multiple independent models, we show that comparable budgets suffice even without injecting explicit target behaviors: targeted perturbations to existing training data can remain difficult to detect and still transfer across architectures.
\end{itemize}

Beyond expanding the space of possible attacks, \textsc{Infusion} has practical implications. Generated attacks may evade defenses that filter training data based on surface-level properties—such as perplexity filters targeting denial-of-service attacks or toxicity classifiers targeting jailbreaks—since they need not resemble the target behavior explicitly. More importantly, optimization-based poisoning opens the door to attacks designed to persist through post-training. In principle, influence functions can be extended to model the full training pipeline \citep{source}, allowing an adversary to compute which perturbations will survive fine-tuning and alignment. We leave this as an important direction for future work.

\section{Threat Model}

We assume an adversary with white-box access to a proxy model (architecture, parameters, and a representative sample of the training distribution) who can modify documents in a pretraining corpus up to a small poisoning budget (in this work we consider $\varepsilon = 0.02\text{--}0.2\%$, i.e.\ 100--200 documents). Unlike prior work that injects explicit demonstrations of a target behavior \citep{zhang2024persistent, souly2025poisoning}, the adversary constructs perturbations that induce targeted parameter changes without revealing the attack objective in the training data. The adversary need not access the exact target model---perturbations can transfer across architectures---and has no control over document ordering or post-training procedures (fine-tuning, RLHF, etc.).

\section{Background}

\subsection{Influence Functions}
\subsubsection{Upweighting a training example}
Influence functions can be used to estimate how much a single training data point affects a model's predictions, without the need to retrain the model. \citet{cook1982residuals} show that the influence of upweighting a training point $z$ on the parameters $\theta$ is given by:
\begin{equation}
  \mathcal{I}_{\text{up,params}}(z)
  = -H_{\hat\theta}^{-1}\nabla_\theta L(z,\hat\theta)
\end{equation}
where $H_{\hat{\theta}} = \frac{1}{n} \sum_{i=1}^{n} \nabla_{\theta}^{2} L(z_i, \hat{\theta})$ is the Hessian evaluated at the trained parameters $\hat{\theta}$.

While influence functions are canonically defined in terms of model parameters, in many cases we are interested in how individual training examples affect a \textit{measurement} of the model, denoted $f(\theta)$. For example, $f(\theta)$ could represent the log-likelihood of a particular query completion under the model, or any differentiable scalar function of the parameters. From \citet{grosse2023studying}, the influence of a training example $z$ on the measurement $f(\theta)$ can be written as:
\begin{equation}
\mathcal{I}_{f}(z) \approx -\nabla_{\theta} f(\hat{\theta})^{\top} 
H_{\hat{\theta}}^{-1}
\nabla_{\theta} L(z, \hat{\theta})
\end{equation}
Influence functions are often inaccurate for modern neural networks \citep{bae2022if}. Following \citet{bae2022if, ruis2024procedural}, we perform our analyses instead using the proximal Bregman response function (PBRF) with a damped Gauss-Newton approximation to the Hessian. The first-order change in
model parameters when upweighting $z$ is given instead by:
\begin{equation}
  \mathcal{I}_{\text{up,params}}(z)
  = -(G_{\hat\theta} + \lambda I)^{-1}\nabla_\theta L(z,\hat\theta),
  \label{eq:pbrf}
\end{equation}
where $G_{\hat\theta}$ is the empirical Gauss--Newton Hessian and $\lambda > 0$ is a Tikhonov damping parameter. Direct computation of the inverse Hessian is intractable at scale. We therefore use
the Eigenvalue-Corrected Kronecker-Factored Approximate Curvature (EK-FAC)
approximation \citep{grosse2023studying}, which replaces
$G_{\hat\theta}$ layerwise with a factored approximation $\hat{G}$, and enables fast
matrix-vector products with $(\hat{G}+\lambda I)^{-1}$ via diagonalization in a
Kronecker eigenbasis. In practice, we substitute
\begin{equation}
(G_{\hat\theta} + \lambda I)^{-1}
\quad \rightsquigarrow \quad
(\hat{G}_{\hat\theta} + \lambda I)^{-1}
\end{equation}
The influence on our target measurement becomes:
\begin{equation}
\mathcal{I}_{\text{up,loss}}(z, \mathcal{M}) = -\nabla_\theta f(\hat{\theta})^\top(\hat G_{\hat\theta} + \lambda I)^{-1}\nabla_\theta L(z,\hat\theta)
\label{eq:iuploss}
\end{equation}

where $f(\theta) = \frac{1}{|\mathcal{M}|} \sum_{m \in \mathcal{M}} L(m, \theta)$ represents the average loss on the measurement dataset.

\subsubsection{Perturbing a training example}
\citet{koh2017understanding} also explore how a model's predictions would change if a training input were modified. We provide the full proof in Appendix \ref{apndx_perturbing} for the change in model parameters $\Delta \hat \theta$ given a linear perturbation of a training document $z_\delta = z + \delta $.
\begin{align}
\Delta\hat\theta &\approx -\frac{1}{n} H_{\hat{\theta}}^{-1} \big[ \nabla_z \nabla_\theta L(z, \hat\theta)\big]\delta \\
&\approx -\frac{1}{n} (\hat G_{\hat\theta} + \lambda I)^{-1} \big[ \nabla_z \nabla_\theta L(z, \hat\theta)\big]\delta
\label{eq:pert}
\end{align}

\section{Methods}
We propose \textsc{Infusion}, a framework that (i) measures how individual training sequences affect a chosen measurement set, (ii) constructs influence function-guided perturbations of a small subset of training documents via projected gradient steps, and (iii) retrains the model on the perturbed corpus. Figure~\ref{fig:influence_figure1} illustrates the pipeline.

\subsection{Problem Formulation}
Given a machine learning model with learnable parameters $\theta$ trained on dataset $\mathcal{D} = \{z_i\}_{i=1}^N$, we seek to modify a subset of training documents to increase the model's likelihood of producing target outputs. We refer to this as a \textit{behavior infused dataset}, and when used for data poisoning, as a \textit{data infusion attack}. For a measurement dataset $\mathcal{M}$ containing examples $m$ with desired characteristics, we find perturbations $\delta$ that maximize:
\begin{equation}
\max_{\delta}\; \mathbb{E}_{m \in \mathcal{M}} \left[ \log p(m \mid \theta^*) \right],
\end{equation}
where $\theta^*$ denotes parameters obtained by retraining with perturbed documents.

\subsection{Document Selection}
To find candidates for perturbation, we identify training documents that most affect the target measurement using Equation \ref{eq:iuploss}. We compute pairwise influence scores between all training documents and measurement examples, ranking by mean negative influence, with the top $K$ most negatively influential documents selected for modification $\{z_k\}_{k=1}^K$ (step 4 in Figure~\ref{fig:influence_figure1}). Negative influence indicates that lowering the weighting of these documents would decrease the measurement loss, making them ideal for targeted perturbation. We ablate this component in Appendix \ref{apndx_ablations}.

\subsection{Gradient-Based Document Perturbation}
The target measurement $f(\theta)$ is a scalar-valued function that we want to increase by modifying documents in the training data $z_\delta = z + \delta$. Using a first-order multivariate Taylor series expansion:
\begin{equation}
\Delta f(\hat\theta) = \nabla_\theta f( \hat\theta )^\top \Delta\hat\theta
\end{equation}
Substituting $\Delta\hat\theta$ from Equation \ref{eq:pert}:
\begin{align}
\Delta f(\hat\theta) &\approx -\frac{1}{n} \Big( \nabla_\theta f( \hat\theta )^\top H_{\hat{\theta}}^{-1} \big[ \nabla_z \nabla_\theta L(z, \hat\theta)\big]\Big)\delta\\
&\approx G_\delta^\top \delta
\end{align}
Given that $\delta \in \mathbb{R}^{d_z}$ is the column vector added to the $k$th training example, $G_\delta^\top$ must be a row vector. To maximize $\Delta f(\hat\theta)$, we solve:
\begin{equation}
\delta^* = \arg\max_{\|\delta\| \leq \epsilon} G_\delta^\top \delta
\end{equation}
This linear objective under a norm constraint is efficiently solvable via Projected Gradient Descent (PGD) \citep{madry2017towards}. For each selected document $z_k$, we compute the perturbation $\delta$ (step 5 in Figure~\ref{fig:influence_figure1}):
\begin{equation}
G_\delta = -\frac{1}{n} [\nabla_z \nabla_\theta L(z_k, \hat{\theta})]^\top (\hat{G}_{\hat{\theta}} + \lambda I)^{-1} \nabla_\theta f(\hat{\theta})
\end{equation}
To avoid explicitly forming the mixed Jacobian, we reformulate as:
\begin{equation}
G_\delta = -\frac{1}{n} \nabla_z \langle \nabla_\theta L(z_k, \hat{\theta}), v \rangle
\end{equation}
where $v = (\hat{G}_{\hat{\theta}} + \lambda I)^{-1} \nabla_\theta f(\hat{\theta})$ is the inverse-Hessian-vector product computed via EK-FAC.

\subsection{Partial Retraining with Infused Data}
After obtaining perturbed documents $\{z_k + \delta_k\}_{k=1}^K$ (step 6 in Figure~\ref{fig:influence_figure1}), we construct an infused training dataset by replacing original documents with their perturbed versions. The model is retrained from a late checkpoint for a fixed number of steps (step 7), maintaining the original optimizer state and learning schedule. Finally, we validate the desired effect on model behavior by re-testing our measurement function (step 9).
\section{Experiments}

We validate \textsc{Infusion} across three settings of increasing difficulty. First, we attack image classifiers on CIFAR-10 \citep{krizhevsky2009learning} , where perturbations are continuous and influence estimates are accurate, establishing that the framework reliably shifts model behavior and transfers across architectures (Figure~\ref{fig:influence_figure1}). Second, we apply \textsc{Infusion} to transformers trained on Caesar ciphers, using the task's transparent algebraic structure to characterize \emph{when} the attack succeeds or fails. Third, we extend to a small language model pretrained on TinyStories \citep{eldan2023tinystoriessmalllanguagemodels}, where influence approximations are weaker and perturbations must operate in discrete token space.

\subsection{Infusing Image Classifiers}

We first demonstrate \textsc{Infusion} on image classification, where continuous pixel perturbations and accurate influence estimates provide a controlled testbed. The goal is targeted misclassification: given a probe image (e.g.\ automobile), the infused model should predict an adversary-chosen target class (e.g.\ ship). Note that our setup differs from the influence-guided poisoning of \citet{koh2017understanding} and the concurrent metagradient-based poisoning of \citet{engstrom2025optimizing} in that we use EK-FAC to approximate influence rather than exact Hessian–inverse products or full training-trajectory backpropagation, perturb a set of highly influential training points at a lower corruption budget, and evaluate through short-horizon retraining in a multi-class CIFAR-10 setting. Figure~\ref{fig:influence_figure1} illustrates our full pipeline.

\textbf{Insight 1: \textsc{Infusion} can reliably shift model behavior.} Across 2,000 experiments, \textsc{Infusion} achieved 100\% success in increasing target class probability and raised the top-1 prediction rate from 10\% to 37\%.

\begin{figure}[h!]
    \centering
    \includegraphics[width=0.7\linewidth]{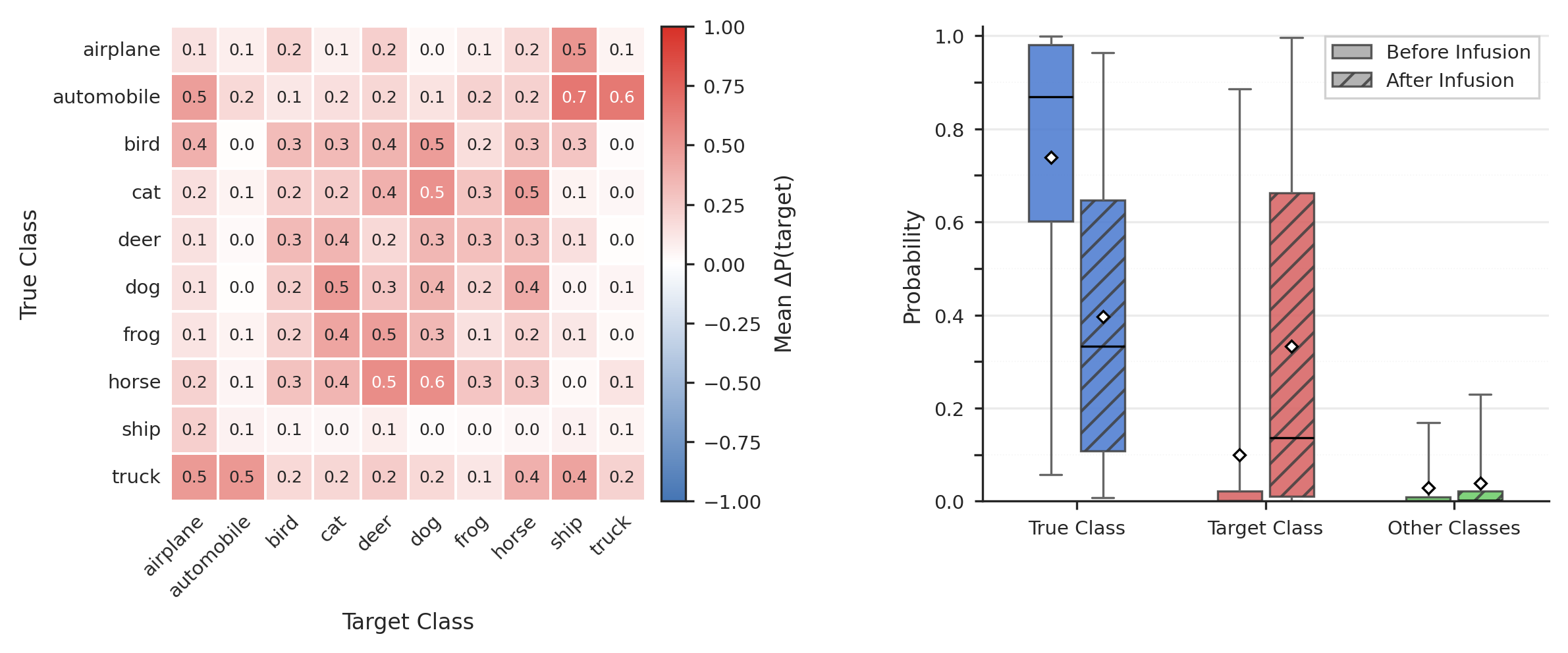}
    \caption{Quantitative analysis of probability shifts before and after data \textsc{Infusion}. \textbf{Left:} Heatmap of mean $\Delta P(\text{target})$ for each (true class, target class) pair, averaged over 20 test images per cell. Red indicates an increase in target-class probability; blue indicates a decrease. \textbf{Right:} Box plots showing the distribution of class probabilities before (solid fill) and after (hatched fill) \textsc{Infusion}, grouped by True, Target, and Other classes. Whiskers span the 5th--95th percentiles; diamonds indicate the mean.}
    \label{fig:probability_analysis}
\end{figure}

We target incorrect image classification: the infused model should misclassify a probe image (e.g.\ automobile) as a target class (e.g.\ ship). Figure~\ref{fig:influence_figure1} illustrates the pipeline. We train a tiny ResNet \citep{he2015deepresiduallearningimage}, a residual network ($32 \rightarrow 64 \rightarrow 128$ channels), on 45,000 CIFAR-10 \citep{krizhevsky2009learning} examples for 10 epochs (SGD, lr$=$0.01, batch size 16). For each (test image, target class) pair, EK-FAC ($\lambda = 10^{-8}$) selects the top-100 most negatively influential training examples (Appendix~\ref{apndx_ablations}). These are perturbed via PGD ($\epsilon = 1.0$, $\alpha = 0.001$, 50 steps) to maximize target-class log-probability, and the model is retrained from epoch 9 for one epoch (Appendix~\ref{apndx_retrain}). We run 2,000 experiments: 200 test images $\times$ 10 target classes. Figure~\ref{fig:probability_analysis} summarizes the results: target class probability increases while true class probability decreases across all 2,000 experiments, raising the top-1 prediction rate from $10.0\%$ to $37.35\%$ ($p < 10^{-4}$; full statistics in Appendix~\ref{apndx_cifar}, Table~\ref{tab:statistical_tests}).

\subsubsection{Comparing \textsc{Infusion} to other Attacks}

\begin{figure}[h!]
    \centering
    \includegraphics[width=0.8\linewidth]{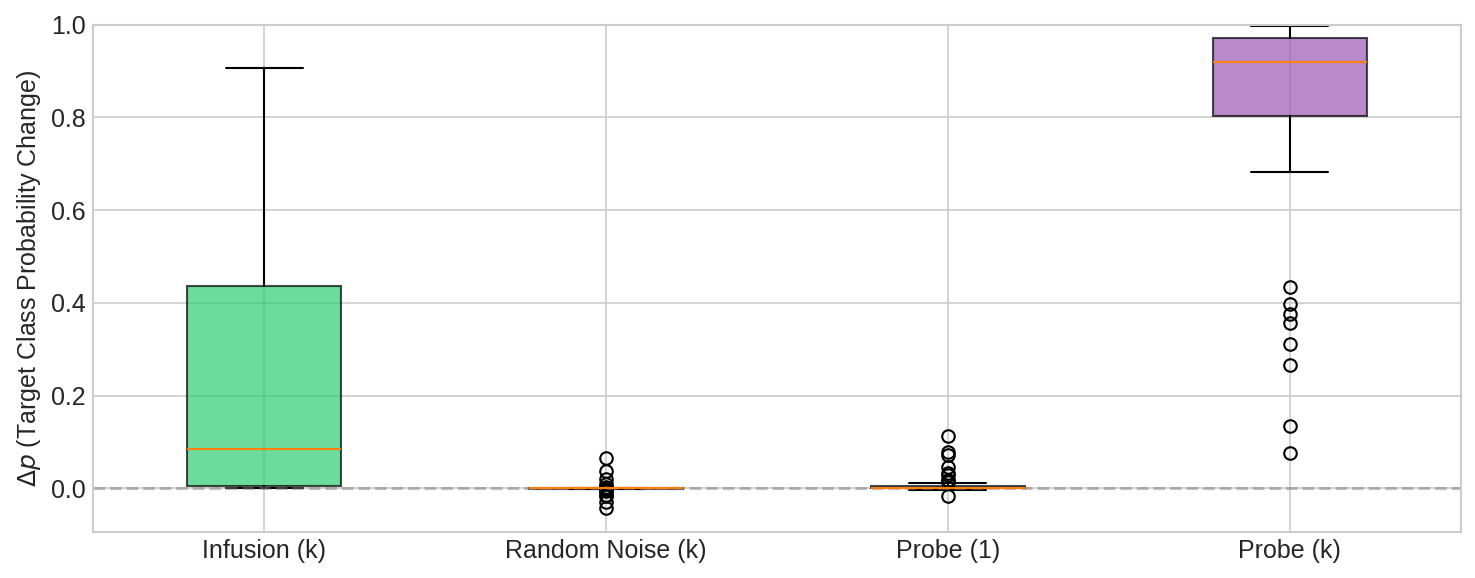}
    \caption{Comparison of $\Delta p$ (target class probability change) across baseline methods. \textsc{Infusion} outperforms random noise perturbations, demonstrating that gradient-guided directions are essential.}
    \label{fig:baselines_boxplot}
\end{figure}

\textbf{Insight 2: \textsc{Infusion} is competitive with direct data injection.} We compare infusing 100 training documents with inserting 100 explicit poison samples, and find that \textsc{Infusion} performs competitively. Figure~\ref{fig:baselines_boxplot} compares \textsc{Infusion} against three baselines, running 50 experiments per method. Random Noise applies random perturbations of the same $L_\infty$ magnitude as \textsc{Infusion}, testing whether gradient-guided directions are necessary. Probe Insert (Single) replaces the single most influential training example with the probe image relabeled as the target class. Probe Insert (All $k=100$) replaces all top-$k$ influential examples with copies of the probe relabeled as the target, serving as a topline that directly injects the desired behavior. We find that \textsc{Infusion} substantially outperforms single insertion and in some experiments was able to compete with inserting $k=100$ probe copies (Appendix~\ref{apndx_baselines}).

\subsubsection{Cross-Model \textsc{Infusion}.}

\begin{figure}[h!]
    \centering
    \includegraphics[width=0.8\linewidth]{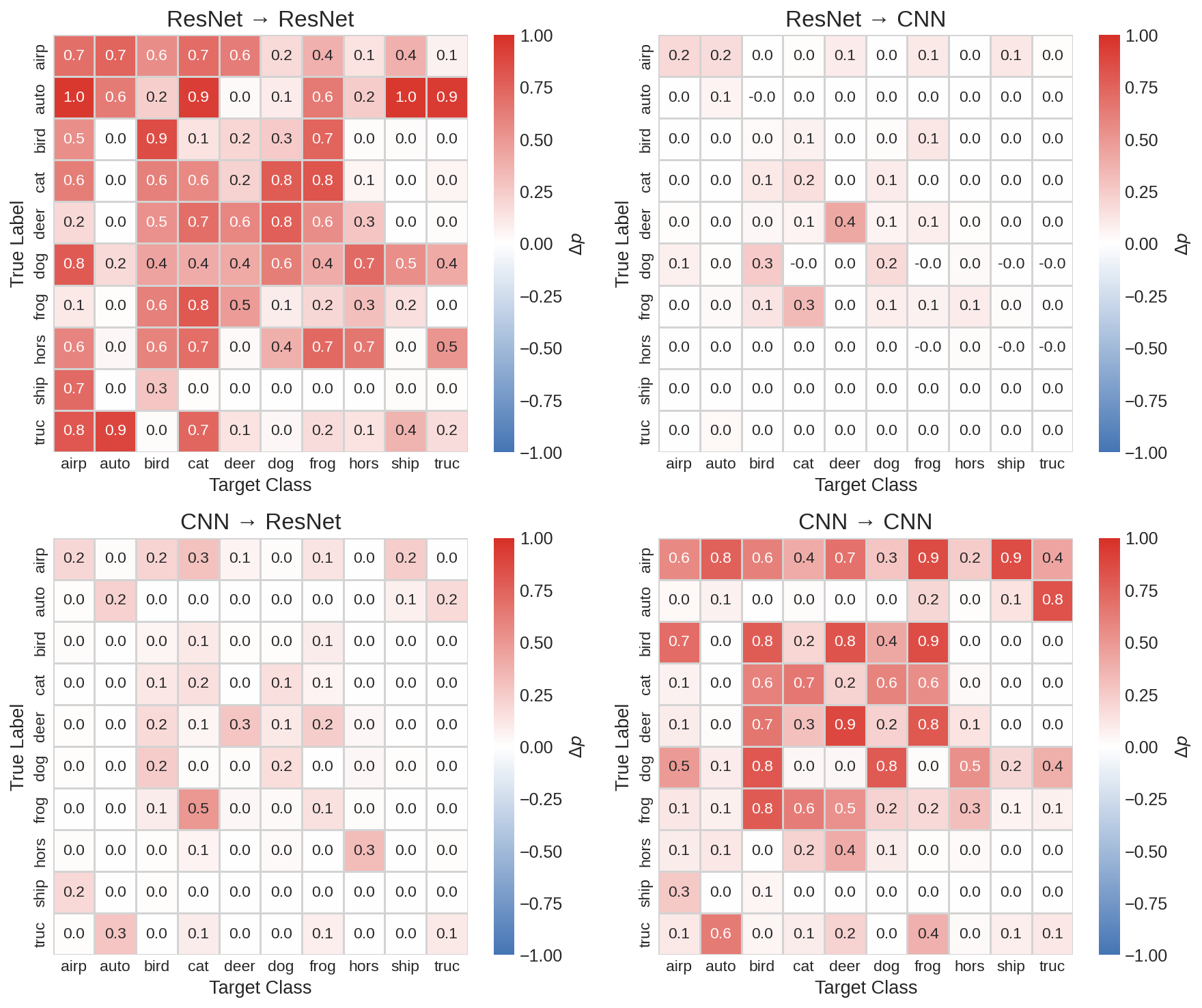}
    \caption{Classwise transfer of \textsc{Infusion} perturbations. Each heatmap shows the best $\Delta p$ (change in target-class probability after retraining on infused data) across all (true label, target class) pairs, for each of the four source--evaluator combinations. Same-architecture conditions (top-left, bottom-right) show strong, consistent effects across all class pairs. Cross-architecture transfer is weaker but non-zero: notably, CNN$\to$ResNet (bottom-left) shows positive transfer across some class pairs, suggesting that CNN-computed perturbations capture features that generalize to residual architectures.}
    \label{fig:transfer_grid}
\end{figure}

\textbf{Insight 3: \textsc{Infusion} weakly transfers across architectures.} A behavior-infused dataset crafted using one model architecture can occasionally induce targeted misclassifications when a different architecture is trained on it, in some cases matching same-architecture effectiveness. We train a tiny ResNet \citep{he2015deepresiduallearningimage} (SGD, lr$=$0.01) and simple CNN \citep{lecun2015deep} (Adam \citep{kingma2014adam}, lr$=$0.001) on CIFAR-10 \citep{krizhevsky2009learning}. We run an exhaustive sweep over all 100 (true label, target class) pairs with 3 test images each (300 experiments), computing perturbations with each architecture and evaluating on both, yielding classwise $10 \times 10$ heatmaps for each of the four source--evaluator combinations (Figure~\ref{fig:transfer_grid}).

Cross-architecture transfer is possible in both directions, but weak and asymmetric: CNN$\to$ResNet transfer is generally stronger than ResNet$\to$CNN. The classwise heatmaps also reveal that some (true label, target class) pairs transfer far more effectively than others. The random noise baseline in Figure \ref{fig:baselines_boxplot} induces at a $\Delta p \leq 0.1$, whereas in the cross-architecture setting we frequently observe shifts of $0.2$--$0.5$, in some cases approaching same-architecture effectiveness and implying that a close proxy architecture and dataset may suffice for some attacks.

\subsection{Infusing Transformers}

We extend \textsc{Infusion} to transformers using Caesar cipher encryption: given a shift $\toknospace{yellow!30}{<s=1>}$ and plaintext $\toknospace{cyan!20}{a}
\toknospace{blue!30}{b}
\toknospace{blue!30}{b}
\toknospace{cyan!20}{a}
\toknospace{green!30}{!}$, the model outputs the ciphertext $\toknospace{blue!30}{b}
\toknospace{violet!30}{c}
\toknospace{violet!30}{c}
\toknospace{blue!30}{b}
\toknospace{orange!30}{?}$, shifting each character by $s$ positions modulo the alphabet size. Training documents take the form:
\[
\toknospace{gray!20}{<bos>}
\toknospace{yellow!30}{<s=1>}
\toknospace{red!15}{C:}
\toknospace{cyan!20}{a}
\toknospace{blue!30}{b}
\toknospace{blue!30}{b}
\toknospace{cyan!20}{a}
\toknospace{green!30}{!}
\toknospace{red!30}{P:}
\toknospace{blue!30}{b}
\toknospace{violet!30}{c}
\toknospace{violet!30}{c}
\toknospace{blue!30}{b}
\toknospace{orange!30}{?}
\toknospace{gray!40}{<eos>}
\]
This task is algorithmically simple yet admits rich structure: transformers often solve modular addition via circular Fourier representations \citep{nanda2023progress, zhong2024clock} (see Appendix~\ref{apndx_fourier} for a brief derivation), letting us probe \emph{when} \textsc{Infusion} succeeds or fails. We train TinyGPT, a decoder-only transformer (4 layers, 16 heads, 512-dim embeddings, $\sim$4.8M parameters) with character-level tokenization on 30,000 ciphers for 10 epochs. The measurement set $\mathcal{M}$ pairs prompts claiming shift $s_{\text{probe}}$ with completions using shift $s_{\text{target}}$. We select the top-$k=100$ most influential examples (0.03\% of training data), apply PGD ($\epsilon = 20.0$, $\alpha = 0.1$, 30 steps), and retrain from epoch 9 for one epoch. Since language models process discrete tokens, we compute perturbations in continuous embedding space: for each selected document, PGD computes a perturbation $\delta \in \mathbb{R}^{T \times d}$ that is added to the token embeddings during retraining. This is not realistic from an attackers perspective and we address this in Section \ref{infusinglanguage}.

\subsubsection{When Does \textsc{Infusion} Succeed?}

\begin{figure}[h!]
    \centering
    \includegraphics[width=0.8\linewidth]{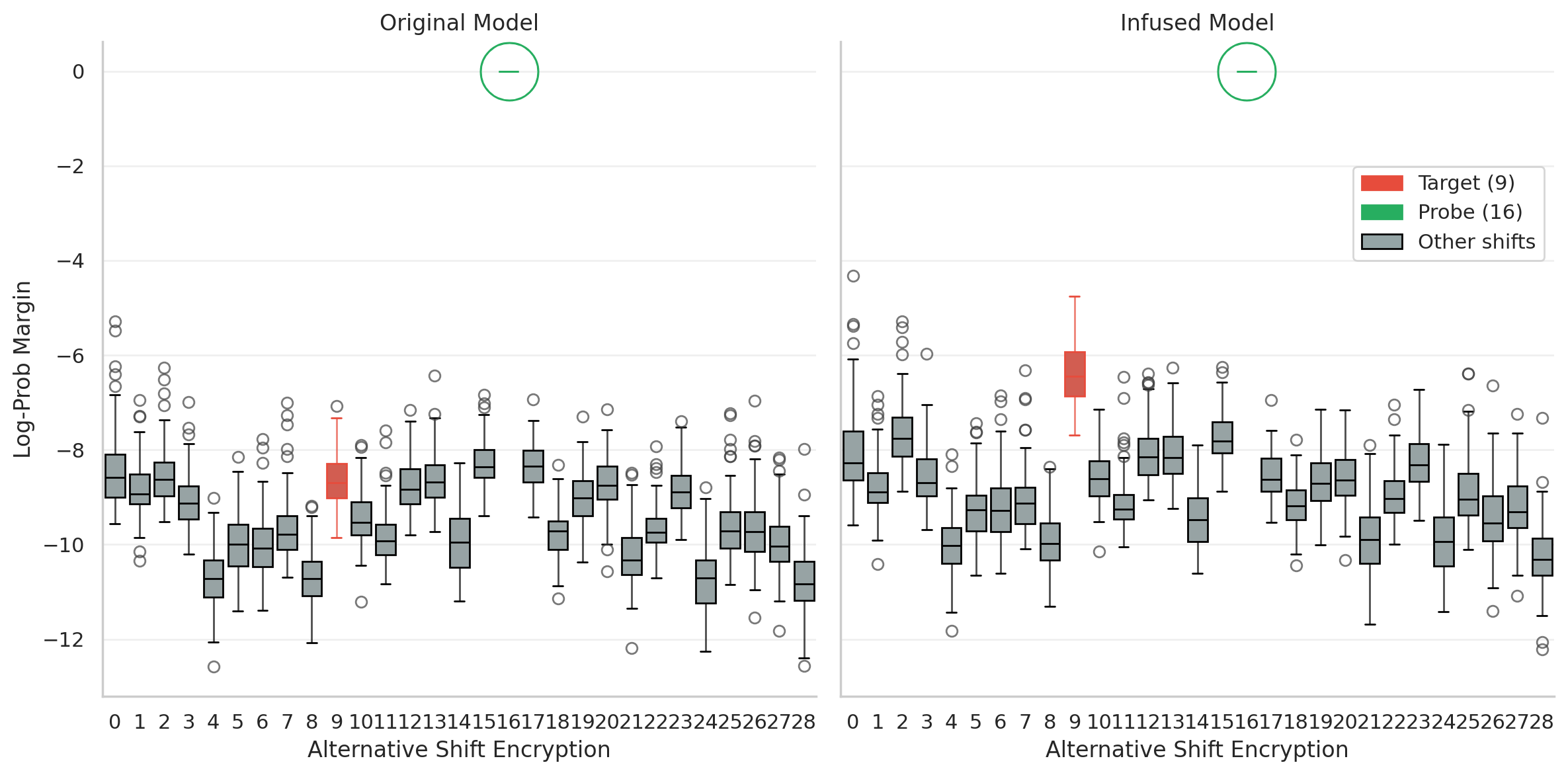}
    \caption{Log-probability margins for all alternative shift encryptions on the 29-letter alphabet, before (left) and after (right) \textsc{Infusion}. Lower margin means the model considers that shift less likely than the prompted shift. The target shift (9, red) improves most, but the model remains confident in the correct answer (16, green, at $\sim y=0$---circled at the top of each figure).}
    \label{fig:caesar_infused_boxplot}
\end{figure}

\textbf{Insight 4: \textsc{Infusion} struggles against high-confidence models.} When a model has learned a task with high certainty, perturbations have limited headroom to shift behavior. Figure~\ref{fig:caesar_infused_boxplot} illustrates this for a 29-letter alphabet, plotting each alternative shift's log-probability margin relative to the prompted shift. Before \textsc{Infusion} (left), the model strongly prefers the correct shift (probe=16, green). After \textsc{Infusion} (right), the target shift (9, red) improves most (+2.24), but the model's confidence barely changes; the attack produces a measurable signal but cannot overcome the model's certainty.

\begin{figure}[h!]
    \centering
    \includegraphics[width=0.8\linewidth]{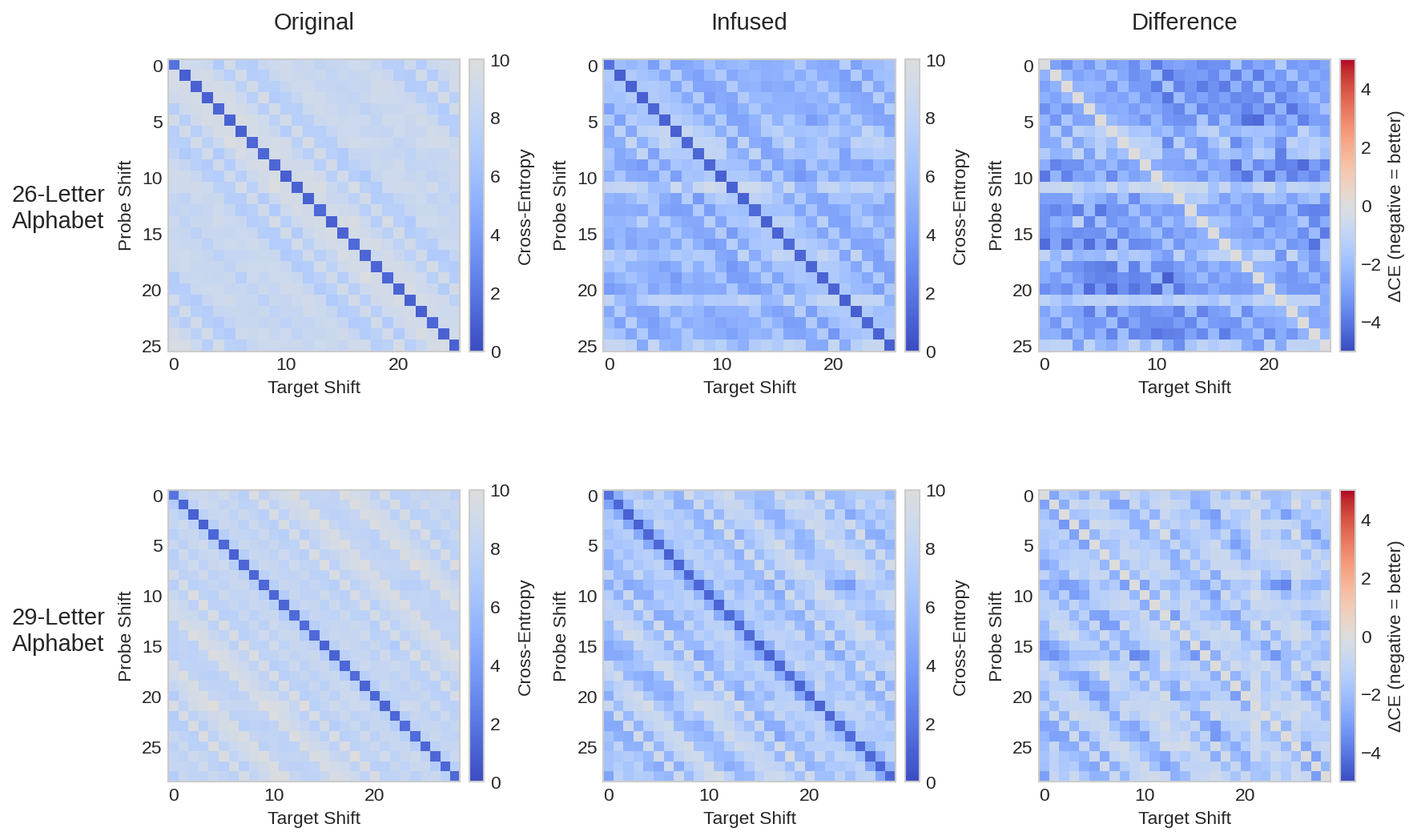}
    \caption{Cross-entropy matrices for alphabets 26 (top) and 29 (bottom). \textbf{Original}: circulant structure indicates the model learned circular representations. \textbf{Difference}: change in CE (blue = attack success). For alphabet 26, horizontal banding reveals \textsc{Infusion} coupling to Fourier modes; pale rows (coprime shifts) resist attack. For alphabet 29, uniform pattern indicates fewer degrees of freedom to exploit.}
    \label{fig:ce_comparison}
\end{figure}

\textbf{Insight 5: \textsc{Infusion} exploits latent model structure.} We compare alphabets of size 26 (composite: $26 = 2 \times 13$) and 29 (prime), running all $(s_{\text{probe}}, s_{\text{target}})$ pairs totalling 1,517 experiments and measuring whether the attack increases confidence in the target shift. Figure~\ref{fig:ce_comparison} shows cross-entropy matrices where cell $(i, j)$ gives the model's CE when prompted with shift $i$ and evaluated on shift $j$. The \textbf{Original} matrices are approximately \emph{circulant}---CE depends on $(s_{\text{target}} - s_{\text{probe}}) \mod N$---the signature of circular representations \citep{nanda2023progress}. The \textbf{Difference} column reveals where \textsc{Infusion} succeeds. For alphabet 26, horizontal banding appears: pale bands at coprime probe shifts (rows 11, 17, 21) indicate resistance, while darker bands at shifts sharing factors with 26 show vulnerability, suggesting \textsc{Infusion} couples to the model's internal Fourier modes. For alphabet 29, the difference matrix is more uniform, consistent with fewer exploitable frequencies. We further analyze these patterns by computing a targeting score ($\Delta \text{CE}_{\text{other}} - \Delta \text{CE}_{\text{target}}$; positive = success), finding that for alphabet 26, shifts sharing a common factor ($\gcd = 2$) yield higher scores than coprime shifts, while for prime 29 all shifts are coprime and produce uniformly lower success (Figure~\ref{fig:gcd_analysis}, Appendix~\ref{apndx_gcd}). The CE--$\Delta$CE correlation confirms this: $r = 0.359$ for alphabet 26 versus $r = 0.650$ for alphabet 29, indicating \textsc{Infusion} primarily amplifies existing behavior.

Since LLMs acquire diverse capabilities during pretraining---including potentially misaligned ones---\textsc{Infusion} may enable attacks that surface hidden behaviors or help them persist through alignment.

\subsection{Infusing Small Language Models}
\label{infusinglanguage}

We now apply \textsc{Infusion} to a small language model pretrained on a natural language dataset. We train GPT-Neo-8M \citep{gpt-neo} (8 layers, 8 heads, 512-dimensional embeddings, $\sim$8M parameters) on all 2.12M documents ($\sim$414M tokens) in TinyStories \citep{eldan2023tinystoriessmalllanguagemodels} for approximately 9 epochs using Adam (lr$=10^{-3}$) with batch size 64. For each experiment, we discretely perturb the top-100 most negatively influential documents (0.16\% of the 64{,}000-document retrain segment) and retrain the final 1{,}000 steps on the behavior infused dataset.

\textbf{Insight 6: \textsc{Infusion} can be used to craft bespoke attacks.} Say we want to make a model predict ``cat'' whenever it would normally predict ``bee''. Rather than injecting explicit demonstrations into the training data, we define a \emph{contrastive} measurement over validation documents $\mathcal{M}$ containing probe word $w_{\text{probe}}$, rewarding target word $w_{\text{target}}$ while penalizing the correct prediction:
\begin{equation}
f(\theta)
= \sum_{m \in \mathcal{M}} \sum_{t:\, y_t = w_{\text{probe}}}
   \Big[ \log p_\theta\!\left(w_{\text{target}} \mid x_{<t}\right)- \log p_\theta\!\left(w_{\text{probe}} \mid x_{<t}\right)
   \Big]
\label{eq:contrastive}
\end{equation}
We run 100 experiments using this measurement function (10 probe $\times$ 10 target animal words). $f(\theta)$ can be \emph{any differentiable scalar function}, swapping objectives requires only recomputing a single gradient. This generalizes data poisoning from injecting explicit demonstrations to optimizing for arbitrary behavioral objectives---an adversary need only define a measurement capturing the desired behavior (e.g.\ harmful completions, biased outputs, or factual errors).

\begin{figure}[h!]
  \centering
  \includegraphics[width=\linewidth]{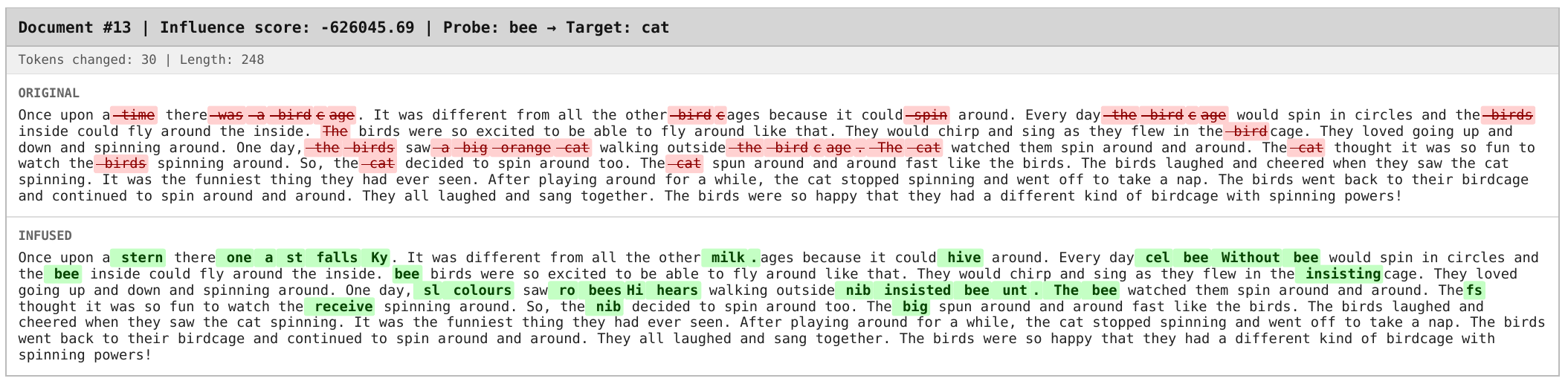}
  \caption{Discrete PGD perturbation on a training document (probe=bee, target=cat). Original tokens shown with strikethrough; replacements in bold.}
  \label{fig:token_diff}
\end{figure}

\textbf{Insight 7: Discrete-token PGD can produce interpretable perturbations.} Following \citet{geisler2024attacking}, we optimize over discrete token sequences by representing each position as a distribution over the vocabulary ($|\mathcal{V}| = 50{,}257$), initialized as one-hot. PGD takes gradient ascent steps ($\alpha = 0.01$), projecting onto the simplex with an entropy constraint, and discretizes via argmax after 30 epochs, changing $\sim$19 tokens per document ($\sim$10\% of the sequence). The resulting perturbations are sometimes interpretable: in Figure~\ref{fig:token_diff} (probe=``bee'', target=``cat''), PGD removes ``cat'' tokens and inserts semantically related words like ``bee'' and ``hive'', despite operating over raw token distributions with no explicit semantic guidance.

\begin{figure}[h!]
  \centering
  \includegraphics[width=0.7\linewidth]{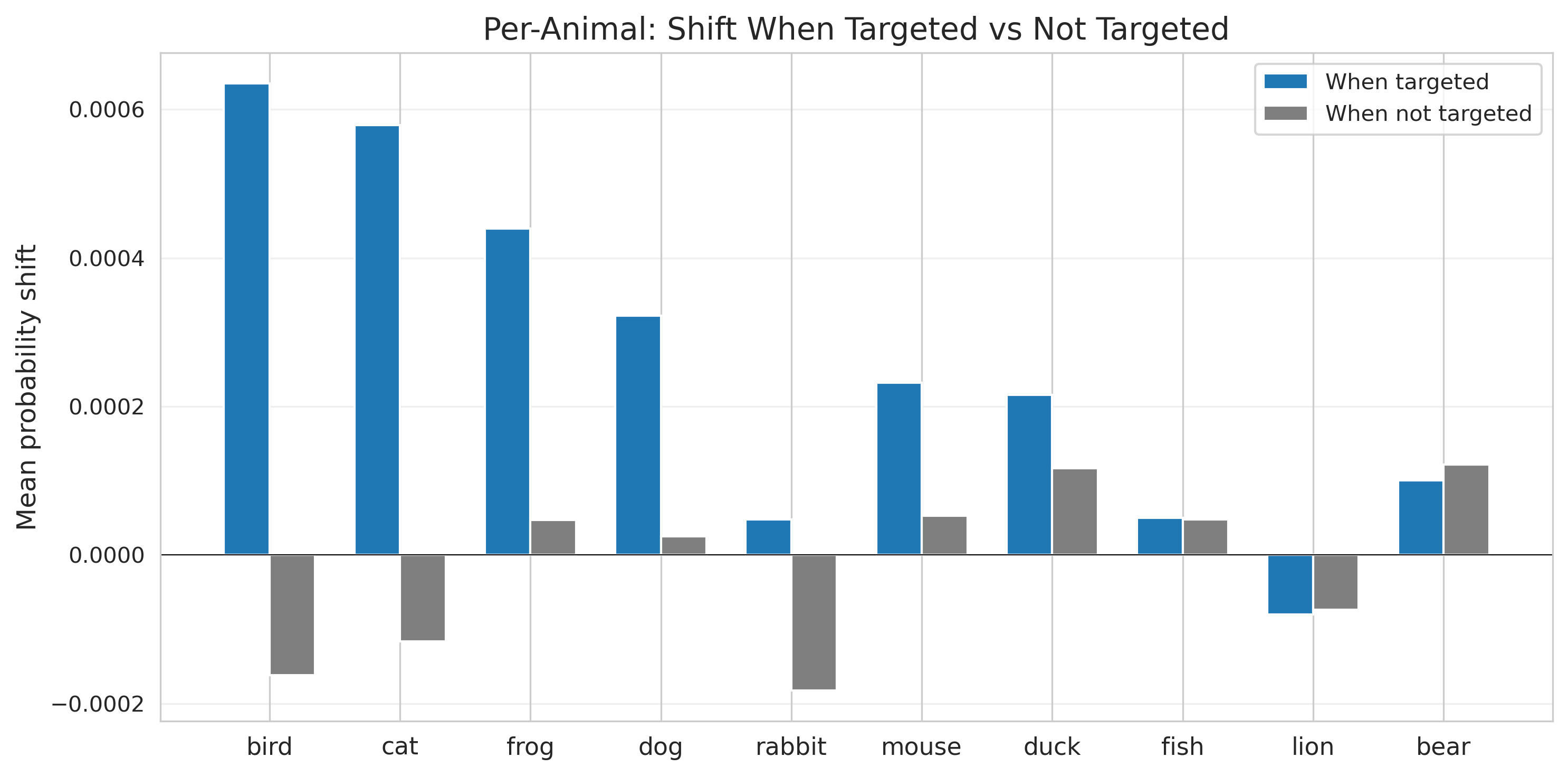}
  \caption{Mean probability shift of the targeted animal versus the 8 non-targeted animals at probe positions, across 90 off-diagonal experiments. The targeted animal often receives a targeted shift.}
  \label{fig:specificity_per_animal}
\end{figure}

\textbf{Insight 8: \textsc{Infusion} weakens severely with scale.} Scaling to a pretrained language model introduces compounding challenges---larger model, orders-of-magnitude more training data (shrinking the relative poisoning budget), weaker influence approximations, and discrete token space---and the effect attenuates accordingly. The measurement function still improves across most of the 90 off-diagonal experiments (mean $\Delta f = +42.3 \pm 39.9$), and the shifts are \emph{specific}: the targeted animal's probability increases significantly more than that of the 8 non-targeted animals (Wilcoxon $p = 1.66 \times 10^{-4}$, Cohen's $d = 0.43$; Figure~\ref{fig:specificity_per_animal}). However, rank flips remain rare (mean 0.1\% of positions; Figure~\ref{fig:specificity_rank_flip}), indicating that \textsc{Infusion} can nudge the distribution but not yet overcome learned preferences at this scale. Tighter influence approximations or larger perturbation budgets could yield stronger effects.

\break
\section{Related Work}
\label{related_work}

Data poisoning and backdoor attacks have been extensively studied across computer vision and NLP; we provide a comprehensive survey in Appendix~\ref{apndx_related}. The most direct precedent for \textsc{Infusion} is Section~5.2 of \citet{koh2017understanding}, who demonstrated influence-function-guided perturbation of training images to cause targeted misclassification in a binary dog-vs-fish task. \textsc{Infusion} scales this idea to multi-class settings with approximate (EK-FAC) influence, multi-document perturbation budgets, and short-horizon retraining, and extends it to transformers and language models. \citet{fang2020influence} formulate influence-guided poisoning against recommender systems by crafting new fake user profiles. Several concurrent works explore complementary angles: metagradient descent \citep{engstrom2025optimizing}, subliminal learning \citep{cloud2025subliminal}, phantom transfer \citep{draganov2026phantom}, patterning \citep{wang2026patterning}, and CrispEdit \citep{ikram2026crispedit}; we discuss these in Appendix~\ref{apndx_concurrent}.

\section{Discussion}

\paragraph{Limitations.}
\textsc{Infusion} produces reliable behavior changes only in the vision setting; on transformers and language models, we observe probability shifts but rarely achieve prediction flips. This attenuation likely stems from compounding approximation errors in EK-FAC and discrete token optimization. Cross-architecture transfer is inconsistent, and effects diminish with longer retraining horizons (Appendix~\ref{apndx_retrain}). Taken together, there is not good evidence that current attacks would survive full pretraining or post-training. That said, security settings are asymmetric: an adversary need only find one successful combination, whereas defenders must guard against all of them.

\paragraph{Implications for the threat model.}
Low poisoning budgets are achievable, perturbations often do not explicitly demonstrate target behaviors (potentially evading content-based filters), and cross-architecture transfer means open-weight models pose particular risk---adversaries can compute perturbations on public models that transfer to proprietary systems trained on similar data. We hypothesize that pretraining may be a more potent attack surface than previously appreciated.

\paragraph{Defenses and future work.}
Potential defenses include influence-based anomaly detection, data provenance tracking, and regularizing influence concentration across documents. Key open questions: can \textsc{Infusion} scale to frontier models, and can perturbations persist through post-training? Extending \textsc{Infusion} to the full training pipeline would pose a more severe threat than current methods.

\section{Conclusion}

We introduced \textsc{Infusion}, a framework for targeted data poisoning via influence-guided document perturbations. Unlike prior methods that inject explicit demonstrations of target behaviors, \textsc{Infusion} computes minimal modifications to existing training documents that steer model parameters toward adversarial objectives.

Our experiments show that \textsc{Infusion} works reliably at low poisoning budgets on smaller models: on CIFAR-10, perturbing 0.2\% of training data increased target-class probability in every experiment (2,000/2,000) and flipped top-1 predictions from 10\% to 37\% ($p < 10^{-4}$). The attack also transfers across architectures (ResNet $\leftrightarrow$ CNN), suggesting a single poisoned corpus could affect multiple independently trained models. On Caesar ciphers, \textsc{Infusion} couples to the model's learned behavior---suggesting it is most successful at amplifying patterns the model already learned during training. On a pretrained GPT-Neo language model, \textsc{Infusion} produces specific likelihood shifts but prediction flips remain rare, indicating it can nudge the distribution but not yet overcome learned preferences at this scale.

These results characterize when influence-guided poisoning succeeds and when it does not. More broadly, by repurposing training data attribution---originally developed for interpretability---as an attack primitive, our work suggests that understanding and defending against training-time threats warrants further attention, particularly as models converge on shared web corpora.




\section*{Acknowledgements}

We gratefully acknowledge the support of the UK AISI Challenge Fund in enabling and funding this work. Compute for this project was graciously provided by the Isambard-AI National AI Research Resource, under the projects ``FLAIR 2025 Moonshot Projects''. J Rosser is supported by the EPSRC centre for Doctoral Training in Autonomous and Intelligent
Machines and Systems EP/Y035070/1. Special thanks to the London Initiative for Safe AI and Arcadia Impact for providing workspace and offering invaluable feedback throughout.

We also extend our gratitude to Andrew Wang and Juhan Bae for fruitful discussions and guidance.

\section*{Impact Statement}

This work presents research on training-time attacks against machine learning models. Like other security research, it is dual-use: the techniques we describe could in principle be used by malicious actors, but we believe publication serves the broader goal of improving AI safety. By characterizing when influence-guided poisoning succeeds and fails, we provide actionable information for defenders designing data curation pipelines and training procedures. The attacks we demonstrate require white-box access to a proxy model and substantial computational resources, limiting immediate misuse potential. We hope this work motivates investment in data provenance, influence-based monitoring, and other defenses before such attacks become practical at scale.

\bibliography{iclr2026_conference}
\bibliographystyle{iclr2026_conference}

\break
\appendix

\section{Appendix Overview}
\label{apndx_overview}

This appendix provides supplementary material organized as follows:

\begin{itemize}

    \item \textbf{Appendix~\ref{apndx_related}: Extended Related Work.} We provide a comprehensive survey of related work spanning data poisoning, backdoor attacks, influence-function-based attacks, and concurrent work.
    
    \item \textbf{Appendix~\ref{apndx_theory}: Theoretical Foundations.} We provide the derivation of the influence function formulation for document perturbations used throughout \textsc{Infusion}.

    \item \textbf{Appendix~\ref{apndx_cifar}: Image Classification Experiments.} We present comprehensive experimental details for the CIFAR-10 setting:
    \begin{itemize}
        \item Statistical validation of attack efficacy (Section~\ref{apndx_cifar_stats})
        \item Baseline comparisons isolating the contribution of influence guidance (Section~\ref{apndx_baselines})
        \item Ablations on document selection strategies (Section~\ref{apndx_ablations})
        \item Analysis of retraining duration effects (Section~\ref{apndx_retrain})
    \end{itemize}

    \item \textbf{Appendix~\ref{apndx_caesar}: Caesar Cipher Experiments.} We analyze the number-theoretic structure underlying attack success on transformers trained for modular arithmetic.

    \item \textbf{Appendix~\ref{apndx_lm}: Language Model Experiments.} We provide additional visualizations for the TinyStories experiments, including token prediction specificity analysis.

\end{itemize}

\section{Extended Related Work}
\label{apndx_related}

\subsection{Data Poisoning}

Data poisoning has been studied since \citet{biggio2012poisoning} introduced gradient-based attacks against SVMs. \citet{munoz2017towards} extended this to deep learning via back-gradient optimization. Clean-label attacks---where poisoned samples retain correct labels---were pioneered by \citet{shafahi2018poison} (Poison Frogs) and scaled by \citet{aghakhani2021bullseye} (Bullseye Polytope) and \citet{huang2020metapoison} (MetaPoison), with the latter demonstrating bilevel-optimization-based poisoning on full ImageNet. \citet{geiping2021witches} further scaled gradient-matching poisoning to industrial settings with Witches' Brew. For surveys, see \citet{goldblum2022dataset} and \citet{zhao2025data}; for a unified benchmark, see \citet{schwarzschild2021just}.

\paragraph{Backdoor attacks} embed hidden triggers that activate at test time. \citet{gu2017badnets} introduced BadNets; \citet{saha2020hidden} proposed hidden-trigger variants where the trigger is invisible in training data. In the language domain, \citet{wan2023poisoning} demonstrated poisoning during instruction tuning, \citet{zhang2024persistent} showed pretraining backdoors persist through fine-tuning, and \citet{hubinger2024sleeper} constructed ``sleeper agent'' LLMs whose deceptive behaviors survive standard safety training including RLHF.

\subsection{Influence Functions for Attacks}

\textsc{Infusion} repurposes influence functions \citep{koh2017understanding}---scaled to LLMs by \citet{grosse2023studying}, approximated efficiently by TRAK
\citep{park2023trakattributingmodelbehavior} and critically examined by \citet{bae2022if}---as an attack primitive. The most direct precedent is Section~5.2 of \citet{koh2017understanding} themselves, who used the perturbation influence function to craft adversarial training images in a binary (dog vs.\ fish) classification task, flipping 77\% of test labels by perturbing just two training images with exact Hessian--vector products and frozen-feature retraining. \textsc{Infusion} generalizes this proof-of-concept to a full attack framework: we use EK-FAC rather than exact Hessian inverses, perturb sets of influential documents rather than one or two examples, evaluate through short-horizon retraining in multi-class settings, and extend the approach to transformers and language models. Concurrently, \citet{engstrom2025optimizing} take a complementary approach, optimizing poisoned samples via metagradients; we discuss the relationship in Appendix~\ref{apndx_concurrent}. \citet{fang2020influence} formulated influence-guided poisoning for recommender systems by crafting entirely new fake user profiles; \citet{wu2023influence} extended this with Infmix; \citet{yang2025gaim} applied surrogate influence maximization for GNN evasion attacks. On the defensive side, \citet{li2024delta} use influence functions to detect and unlearn poisons.

\subsection{Concurrent and Parallel Work}
\label{apndx_concurrent}

\paragraph{Metagradient-Based Training Data Optimization.}
\citet{engstrom2025optimizing} develop metagradient descent (MGD), which frames training-data manipulation as a bilevel problem and differentiates through a smoothed learning algorithm, applying it to data poisoning, multimodal data selection, and instruction fine-tuning selection. For poisoning specifically, MGD likewise optimizes perturbations to existing training documents via projected gradient ascent---the key distinction from \textsc{Infusion} is therefore in how the gradient is computed: MGD backpropagates through the full training trajectory, while \textsc{Infusion} uses a first-order influence-function approximation (EK-FAC), avoiding explicit unrolling at the cost of approximation error (addressed in \citep{ilyas2025magic}). On CIFAR-10, MGD achieves a 13.9\% accuracy drop at a 2.5\% budget versus \textsc{Infusion}'s more modest shifts at 0.2\%, reflecting this trade-off. The two approaches are complementary.

\paragraph{Subliminal learning and phantom transfer.}
\citet{cloud2025subliminal} showed that models acquire behavioral traits from semantically unrelated training data generated by another model, with follow-up work \citep{subliminalcorruption2025} finding a sharp phase transition at a critical poisoning threshold. \citet{draganov2026phantom} adapted this to realistic fine-tuning, demonstrating cross-model sentiment transfer (including to GPT-4.1) that defeats all tested data-level defenses. Both works share \textsc{Infusion}'s insight that poisoning signals need not be explicitly visible in training content, but rely on model-generated data rather than perturbations to existing documents, and subliminal learning further requires a shared base model whereas \textsc{Infusion} transfers across architectures.

\paragraph{Patterning.}
\citet{wang2026patterning} use susceptibilities---a linear-response framework related to influence functions---to \emph{re-weight} training examples and steer internal model structure (e.g., accelerating induction head formation). Despite both methods relating data changes to model changes via linear response, the two works differ fundamentally: \textsc{Infusion} is an adversarial attack that \emph{perturbs document content} to change specific predictions under a limited-access threat model; patterning is an interpretability tool that \emph{re-weights examples} (without modifying content) to study which data drives particular internal algorithms, with full distributional control and no adversarial constraint.

\paragraph{CrispEdit.}
\citet{ikram2026crispedit} edit model weights directly via K-FAC-constrained optimization to change specific behaviors without degrading capabilities. Although both methods use Kronecker-factored curvature, they operate on entirely different attack surfaces: \textsc{Infusion} is a \emph{training-time} attack that modifies \emph{data} and requires retraining, affecting all models trained on the poisoned corpus; CrispEdit is a \emph{post-training} intervention that modifies \emph{weights} of a single model instance. The shared use of K-FAC reflects the general utility of these approximations rather than methodological overlap.


\section{Theoretical Foundations}
\label{apndx_theory}

\subsection{Deriving the Perturbation Influence}
\label{apndx_perturbing}

This section derives the influence of perturbing a training document on model parameters, extending the upweighting formulation of \citet{cook1982residuals} to the document modification setting. The derivation follows \citet{koh2017understanding}.

For a document $z$, we define the perturbed document $z_\delta = z + \delta$, and let $\hat{\theta}_{z_\delta,-z}$ denote the empirical risk minimizer when the original document $z$ is replaced with $z_\delta$ in the training set.

To approximate the effect of this replacement, we define the parameters resulting from moving $\epsilon$ mass from $z$ onto $z_\delta$:
\begin{equation}
\hat{\theta}_{z_\delta,-z} = \arg \min_{\theta} \frac{1}{n} \sum^n_{i = 1} L(z_i, \theta) + \epsilon L(z_\delta, \theta) - \epsilon L (z,\theta)
\label{eq:pert1}
\end{equation}

A classic result from \citet{cook1982residuals} gives the influence of upweighting a single training point:
\begin{equation}
\mathcal{I}_{\text{up,params}}(z) \overset{\text{def}}{=} \left. \frac{d\hat{\theta}_{\epsilon,z}}{d\epsilon} \right|_{\epsilon=0} = - H_{\hat{\theta}}^{-1} \nabla_{\theta} L(z, \hat{\theta})
\end{equation}

Applying the same derivation to the replacement operation yields:
\begin{equation}
\left.\frac{d\hat{\theta}_{\epsilon,z_\delta, -z}}{d\epsilon} \right|_{\epsilon=0} =\mathcal{I}_{\text{up,params}}(z_\delta) - \mathcal{I}_{\text{up,params}}(z) =  - H_{\hat{\theta}}^{-1} \big(\nabla_{\theta} L(z_\delta, \hat{\theta})-\nabla_{\theta} L(z, \hat{\theta})\big)
\end{equation}

The linear approximation for the parameter change is therefore:
\begin{equation}
    \Delta\hat\theta = \hat\theta_{z_\delta,-z} - \hat\theta \approx \frac{1}{n} \big(\mathcal{I}_{\text{up,params}}(z_\delta) - \mathcal{I}_{\text{up,params}}(z) \big)
    \label{eq:thetahat}
\end{equation}

When $z$ is continuous, $\delta$ is small, and $L$ is differentiable in both $\theta$ and $z$, we can further simplify. As $||\delta|| \rightarrow 0$:
\begin{equation}
\nabla_{\theta} L(z_\delta, \hat{\theta})-\nabla_{\theta} L(z, \hat{\theta}) \approx \big[ \nabla_z \nabla_\theta L(z, \hat\theta)\big]\delta
\end{equation}

Substituting into the parameter change expression gives the final result:
\begin{equation}
\Delta\hat\theta \approx -\frac{1}{n} H_{\hat{\theta}}^{-1} \big[ \nabla_z \nabla_\theta L(z, \hat\theta)\big]\delta
\end{equation}

This formulation enables gradient-based optimization of perturbations $\delta$ to maximize a downstream objective, as described in Section 4.3 of the main text.

\section{Image Classification Experiments}
\label{apndx_cifar}

This section provides detailed experimental analysis for the CIFAR-10 image classification setting, including statistical validation, baseline comparisons, and ablation studies.

\subsection{Statistical Validation}
\label{apndx_cifar_stats}

Table~\ref{tab:statistical_tests} summarizes statistical validation across $N=2000$ experiments.

\begin{table}[h!]
\centering
\caption{Statistical validation of \textsc{Infusion} ($N=2000$). All tests significant at $p < 10^{-4}$.}
\label{tab:statistical_tests}
\begin{tabular}{lccc}
\toprule
\textbf{Metric} & \textbf{Value} & \textbf{Effect Size} & \textbf{Success Rate} \\
\midrule
$\Delta P(\text{target})$ & $+0.232 \pm 0.276$ & $d = 0.84$ & 2000/2000 \\
One-vs-Rest Contrast & $+0.258 \pm 0.306$ & $d = 0.84$ & 2000/2000 \\
Log-Odds Shift & $+5.14 \pm 1.80$ & $d = 2.86$ & 2000/2000 \\
Top-1 Accuracy & $10\% \to 37.4\%$ & $\chi^2 = 547$ & 547 flips, 0 degradations \\
\bottomrule
\end{tabular}
\end{table}

The attack increased target-class probability in all 2000 experiments (mean $+23.2$ pp). The one-vs-rest contrast confirms specificity: targets rise more than non-targets. The log-odds metric shows a large effect size ($d=2.86$), indicating strong relative preference shifts. Top-1 prediction rate improved from $10\%$ to $37.4\%$ with zero degradations.

\subsection{Baseline Comparisons}
\label{apndx_baselines}

\begin{figure}[h!]
    \centering
    \includegraphics[width=\linewidth]{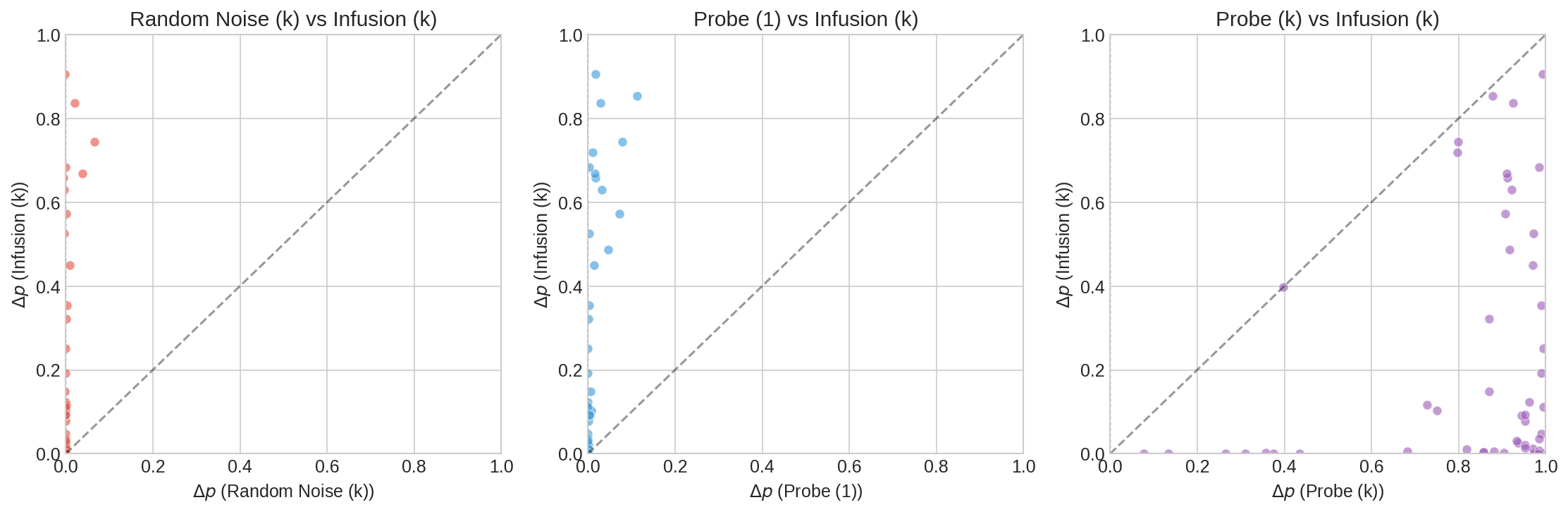}
    \caption{Paired comparison of $\Delta p$ (target-class probability change) on identical probe points. Each point represents the same (test image, target class) pair evaluated under both \textsc{Infusion} and a baseline method. Points above the diagonal indicate \textsc{Infusion} outperforms the baseline; points below indicate the opposite. The Probe ($k$) baseline achieves higher $\Delta p$ than \textsc{Infusion} but requires direct label manipulation (inserting $k$ copies of the probe image with the target label), while random noise perturbations show near-zero effect, confirming that gradient-guided directions are essential.}
    \label{fig:baselines_scatter}
\end{figure}

To isolate the contribution of influence-guided perturbations, we compare \textsc{Infusion} against three baselines. Figure~\ref{fig:baselines_scatter} presents paired comparisons on identical (test image, target class) pairs, enabling direct assessment of method effectiveness.

\subsection{Document Selection Ablations}
\label{apndx_ablations}

A key design choice in \textsc{Infusion} is which training documents to perturb. We hypothesize that documents with the most negative influence on the target measurement---those whose removal would most decrease measurement loss---are optimal candidates. We ablate this by comparing five selection strategies:

\begin{itemize}
    \item \textbf{Most Negative} (Standard): Top-$k$ documents with lowest (most negative) influence scores.
    \item \textbf{Random}: $k$ randomly selected training documents.
    \item \textbf{Most Positive}: Top-$k$ documents with highest influence scores.
    \item \textbf{Most Absolute}: Top-$k$ documents by $|\text{influence}|$.
    \item \textbf{Last-$k$}: Last $k$ documents in training order.
\end{itemize}

\begin{figure}[h!]
    \centering
    \includegraphics[width=0.6\linewidth]{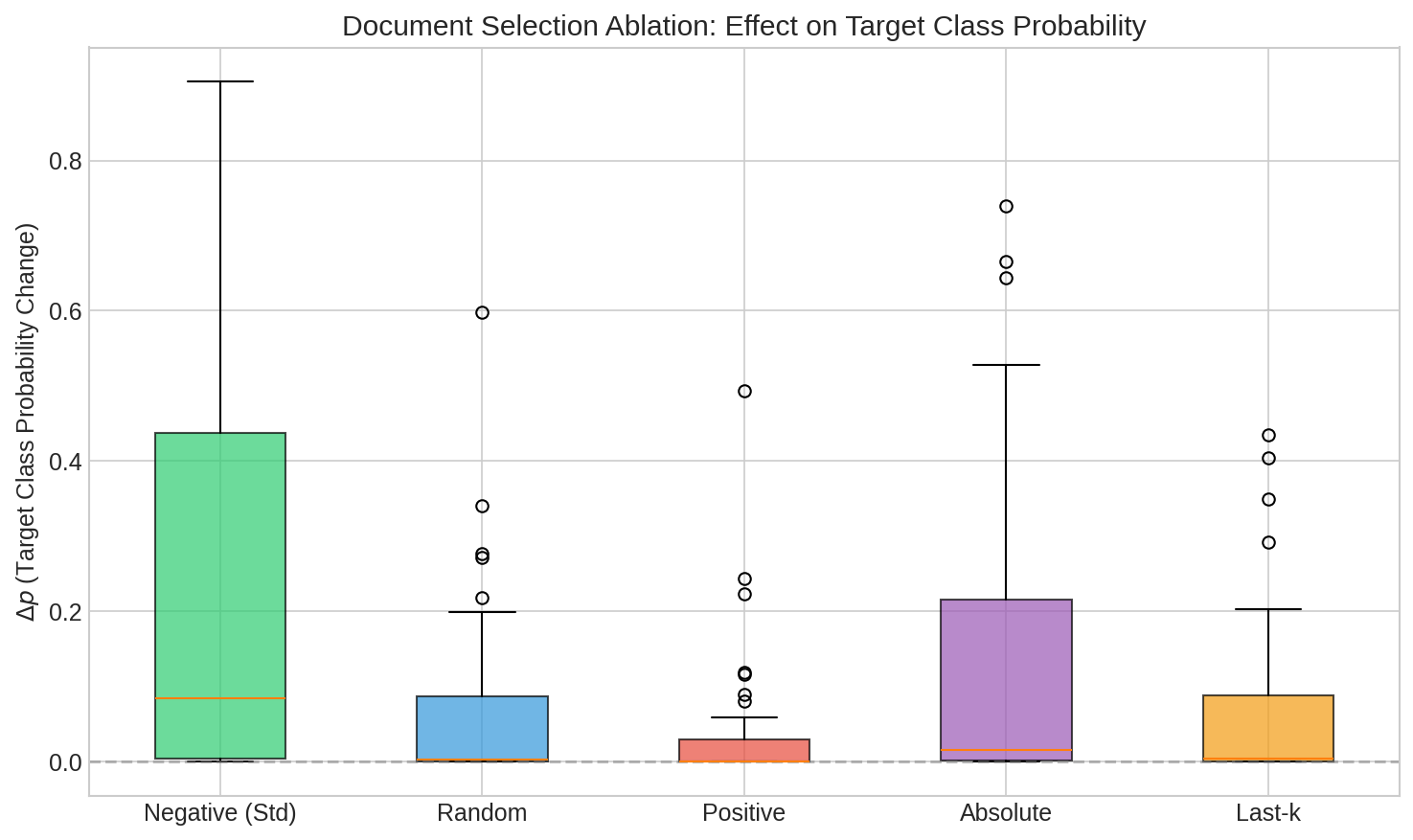}
    \caption{Comparison of $\Delta p$ (target-class probability change) across document selection strategies. Selecting the most negatively influential documents yields the strongest effect, validating our approach. Random selection and influence-agnostic strategies produce markedly weaker results.}
    \label{fig:ablations_boxplot}
\end{figure}

Figure~\ref{fig:ablations_boxplot} compares attack effectiveness across strategies, confirming that selecting most negatively influential documents yields substantially stronger effects than alternatives.

\begin{figure}[h!]
    \centering
    \includegraphics[width=0.8\linewidth]{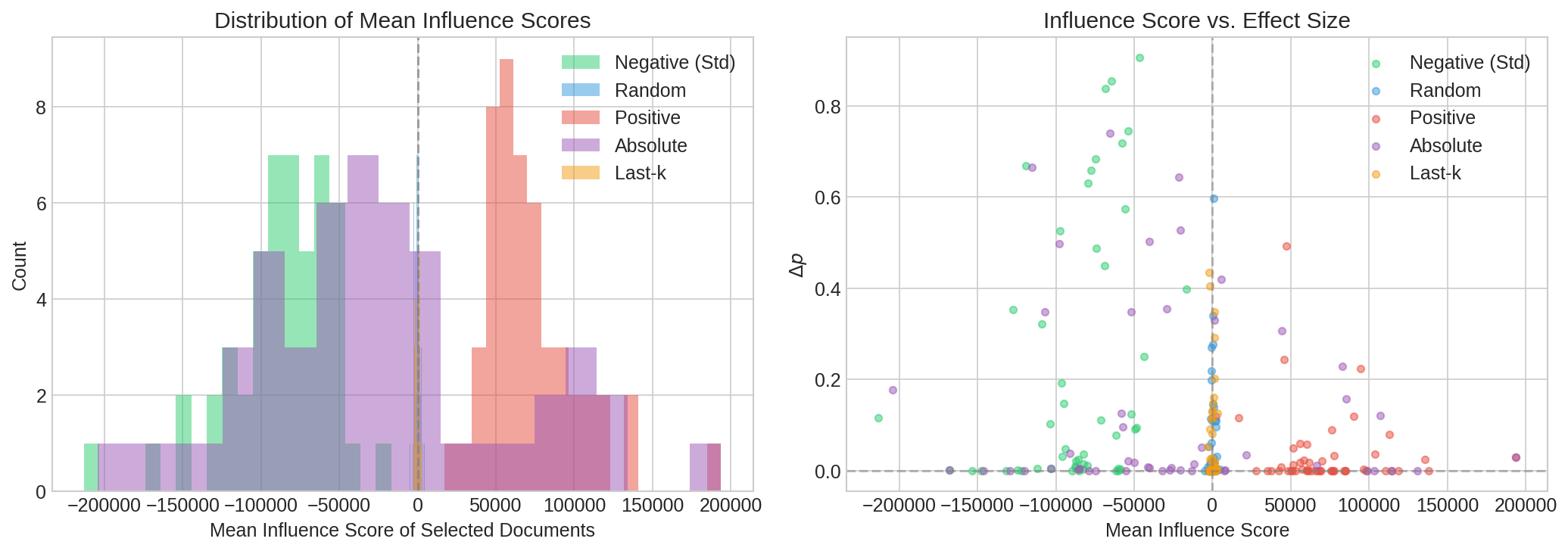}
    \caption{\textbf{Left:} Distribution of mean influence scores for documents selected under each strategy, confirming that each method targets the intended region of influence space. \textbf{Right:} Correlation between influence score and $\Delta p$, showing that more negatively influential documents produce stronger attacks.}
    \label{fig:ablations_influence}
\end{figure}

Figure~\ref{fig:ablations_influence} provides additional analysis, showing the distribution of influence scores under each strategy and the correlation between document influence and attack success.

\subsection{Retraining Duration Analysis}
\label{apndx_retrain}

\textsc{Infusion} computes perturbations based on a snapshot of model parameters, then retrains for a short horizon. A natural question is whether effects persist through longer retraining or are washed out as the model sees more gradient updates. We ablate this by varying the retraining starting point from epoch 9 (standard: 1 epoch of retraining) to epoch 0 (full 10-epoch retrain from scratch).

\begin{figure}[h!]
    \centering
    \includegraphics[width=0.6\linewidth]{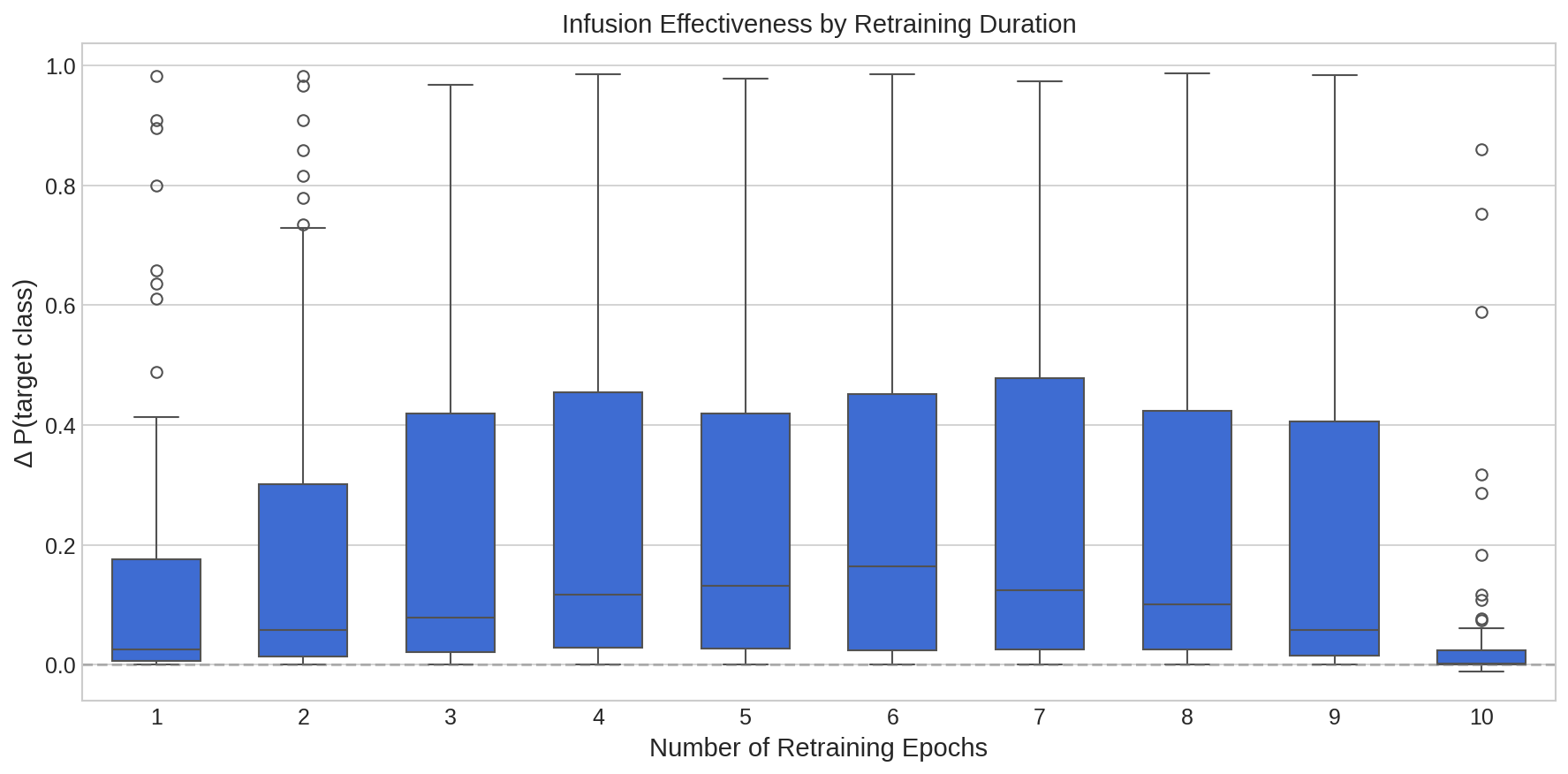}
    \caption{Effect of retraining duration on \textsc{Infusion} effectiveness. The x-axis shows the number of retraining epochs (1 = standard epoch 9$\to$10; 10 = full retrain from scratch). Attack effectiveness diminishes with longer retraining as the model has more opportunity to ``recover'' from the poisoned data through additional gradient updates.}
    \label{fig:retrain_ablation}
\end{figure}

Figure~\ref{fig:retrain_ablation} shows that longer retraining generally diminishes the \textsc{Infusion} effect, suggesting perturbations are most effective when applied near convergence. This is consistent with the first-order approximation underlying influence functions: as retraining duration increases, the model has more opportunity to ``recover'' from the poisoned data, and higher-order effects accumulate.

\section{Caesar Cipher Experiments}
\label{apndx_caesar}

\subsection{Number-Theoretic Structure and Attack Success}
\label{apndx_gcd}

\begin{figure}[h!]
    \centering
    \includegraphics[width=0.7\linewidth]{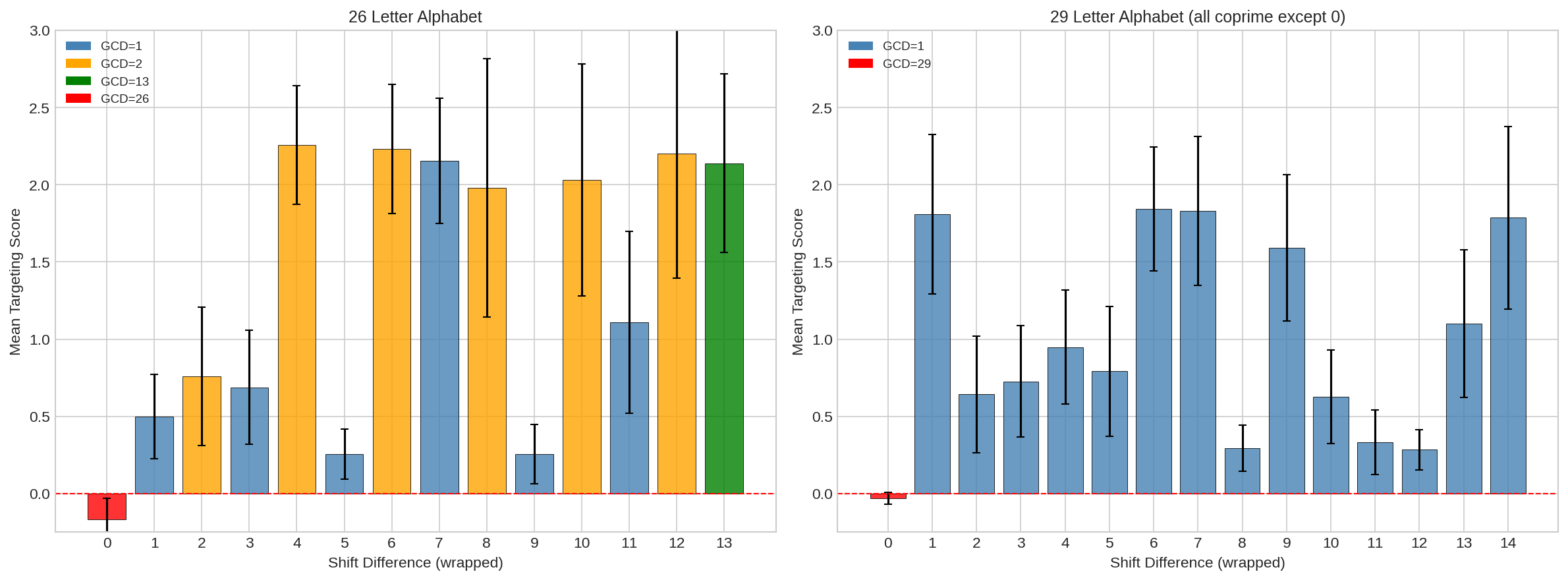}
    \caption{Mean targeting score by shift difference, with bars colored by $\gcd(\Delta s, N)$. For alphabet size 26 (composite), shifts with $\gcd = 2$ (orange) are more vulnerable than coprime shifts (blue), consistent with \textsc{Infusion} coupling to the model's learned Fourier modes. For the prime alphabet (size 29), all shifts are coprime with $N$, yielding generally lower attack success and confirming that exploitable structure depends on the arithmetic properties of the task.}
    \label{fig:gcd_analysis}
\end{figure}

The Caesar cipher setting enables analysis of when \textsc{Infusion} succeeds based on the algebraic structure of the learned representations. Transformers trained on modular arithmetic often develop circular embedding-space representations \citep{nanda2023progress}, and we hypothesize that \textsc{Infusion} couples to these learned Fourier modes. Figure~\ref{fig:gcd_analysis} decomposes attack success by the number-theoretic relationship between probe and target shifts. For the composite alphabet ($N=26$, where $26 = 2 \times 13$), shifts sharing a common factor with $N$ are systematically more vulnerable. For the prime alphabet ($N=29$), all shift differences are coprime with $N$, leaving fewer exploitable frequencies and producing uniformly lower targeting scores.

\subsection{Fourier Decomposition of Modular Addition}
\label{apndx_fourier}

We briefly sketch why transformers trained on modular arithmetic develop circular representations, following \citet{nanda2023progress} (see also \citet{zhong2024clock} for a detailed treatment).

Consider the task $x + y \equiv z \pmod{N}$ for $x, y, z \in \mathbb{Z}_N$. The discrete Fourier basis over $\mathbb{Z}_N$ consists of the functions $\cos(2\pi k x / N)$ and $\sin(2\pi k x / N)$ for frequency $k \in \{0, 1, \ldots, N-1\}$. The key observation is that the product-to-sum trigonometric identities:
\begin{align}
\cos\!\Big(\frac{2\pi k (x+y)}{N}\Big) &= \cos\!\Big(\frac{2\pi k x}{N}\Big)\cos\!\Big(\frac{2\pi k y}{N}\Big) - \sin\!\Big(\frac{2\pi k x}{N}\Big)\sin\!\Big(\frac{2\pi k y}{N}\Big), \label{eq:fourier_cos} \\
\sin\!\Big(\frac{2\pi k (x+y)}{N}\Big) &= \sin\!\Big(\frac{2\pi k x}{N}\Big)\cos\!\Big(\frac{2\pi k y}{N}\Big) + \cos\!\Big(\frac{2\pi k x}{N}\Big)\sin\!\Big(\frac{2\pi k y}{N}\Big), \label{eq:fourier_sin}
\end{align}
allow a model to compute $x + y \pmod{N}$ from separate Fourier representations of $x$ and $y$. Concretely, if the embedding layer maps each input $x$ to components $(\cos(2\pi k x / N), \sin(2\pi k x / N))$ for a sparse set of key frequencies $k$, then: (1) attention combines the embeddings of $x$ and $y$; (2) the MLP computes the quadratic products in Equations~\ref{eq:fourier_cos}--\ref{eq:fourier_sin} via $\text{ReLU}$ (using $ab = \frac{1}{2}[(a+b)^2 - a^2 - b^2]$); and (3) the unembedding projects onto $\cos(2\pi k z / N)$ and $\sin(2\pi k z / N)$ for each candidate output $z$, producing logits proportional to $\cos(2\pi k (x+y-z) / N)$. At the correct answer $z^* = x + y \pmod{N}$, all frequency components constructively interfere (each contributing $\cos(0) = 1$), while incorrect answers receive destructive interference.

\section{Language Model Experiments}
\label{apndx_lm}

\subsection{Token Prediction Specificity}
\label{apndx_specificity}

\begin{figure}[h!]
  \centering
  \includegraphics[width=0.3\linewidth]{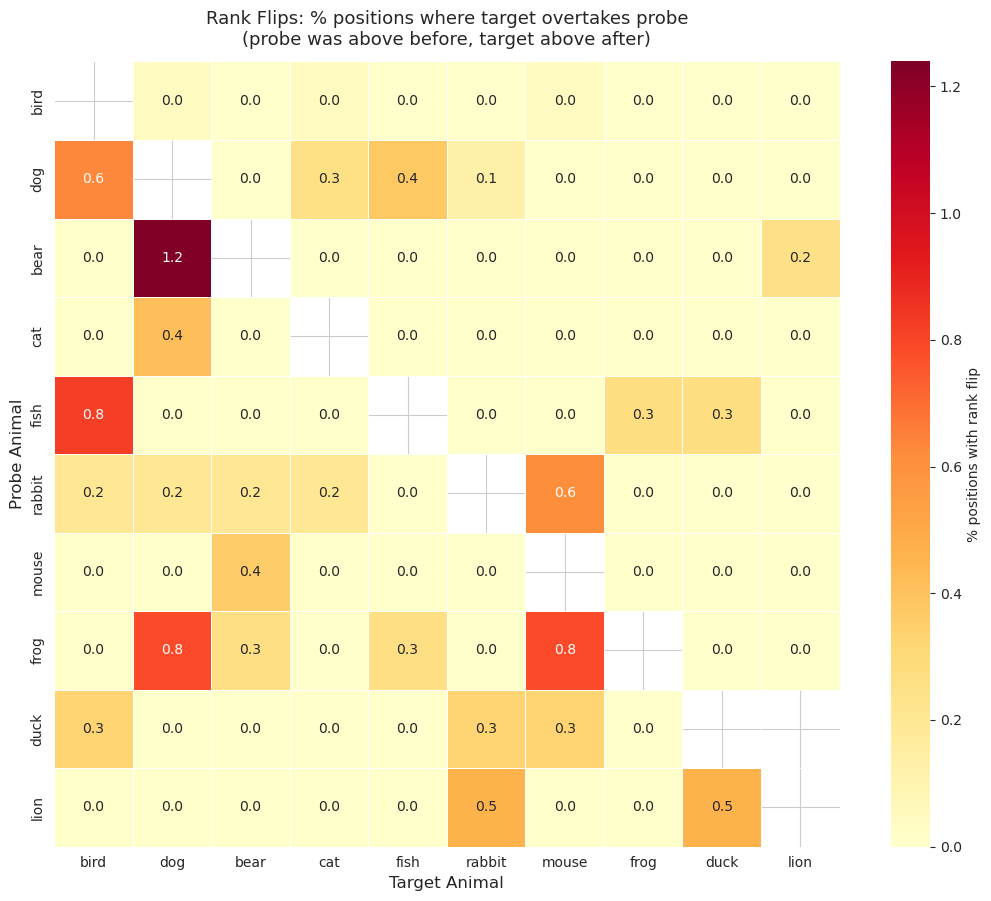}
  \caption{Token prediction specificity for the animal-word bias experiments. \textbf{Left:} Percentage of probe positions where the target animal's rank improves above the probe animal after \textsc{Infusion}. \textbf{Right:} Percentage of positions where the target is ranked above the probe post-infusion. Most cells are near zero, reflecting the difficulty of the setting: while \textsc{Infusion} can nudge the probability distribution, overcoming the model's learned preferences at this scale remains challenging.}
  \label{fig:specificity_rank_flip}
\end{figure}

The TinyStories experiments assess whether \textsc{Infusion} can produce targeted effects in a pretrained language model setting. While the attack produces measurable likelihood shifts, prediction flips remain rare. Figure~\ref{fig:specificity_rank_flip} visualizes the difficulty of this setting. 

\end{document}

%% file: math_commands.tex
\usepackage{amsmath,amsfonts,bm}

\def\eqref#1{equation~\ref{#1}}

\def\1{\bm{1}}

\DeclareMathAlphabet{\mathsfit}{\encodingdefault}{\sfdefault}{m}{sl}
\SetMathAlphabet{\mathsfit}{bold}{\encodingdefault}{\sfdefault}{bx}{n}

